\pdfoutput=1

\documentclass[11pt]{article}

\usepackage{ACL2023}
\usepackage{times}
\usepackage{latexsym}
\usepackage{amsmath}
\usepackage{graphicx}
\usepackage{multirow}

\usepackage[T1]{fontenc}

\usepackage[utf8]{inputenc}

\usepackage{microtype}

\usepackage{inconsolata}

\title{Consolidating Strategies for Countering Hate Speech Using Persuasive Dialogues}

\author{Sougata Saha and Rohini Srihari\\
State University of New York at Buffalo\\
Department of Computer Science and Engineering\\
\texttt{\{sougatas, rohini\}@buffalo.edu}}

\begin{document}
\maketitle
\begin{abstract}
Hateful comments are prevalent on social media platforms. Although tools for automatically detecting, flagging, and blocking such false, offensive, and harmful content online have lately matured, such reactive and brute force methods alone provide short-term and superficial remedies while the perpetrators persist. With the public availability of large language models which can generate articulate synthetic and engaging content at scale, there are concerns about the rapid growth of dissemination of such malicious content on the web. There is now a need to focus on deeper, long-term solutions that involve engaging with the human perpetrator behind the source of the content to change their viewpoint or at least bring down the rhetoric using persuasive means. To do that, we propose defining and experimenting with controllable strategies for generating counter-arguments to hateful comments in online conversations. We experiment with controlling response generation using features based on (i) argument structure and reasoning-based Walton argument schemes, (ii) counter-argument speech acts, and (iii) human characteristics-based qualities such as Big-5 personality traits and human values. Using automatic and human evaluations, we determine the best combination of features that generate fluent, argumentative, and logically sound arguments for countering hate. We further share the developed computational models for automatically annotating text with such features, and a silver-standard annotated version of an existing hate speech dialog corpora.\footnote{OFFENSIVE CONTENT WARNING: This report contains some examples of hateful content, which could be offensive and cause you distress. It is strictly for the purposes of enabling this research, and we have sought to minimize the number of examples where possible. Please contact the authors for code.}


\end{abstract}

\section{Introduction}

Hateful comments are prevalent on social media platforms \cite{schmidt-wiegand-2017-survey} and often go unchecked. While most available solutions to tackling online hate resort to brute force methods such as content flagging and blocking, by not addressing the underlying cause, such methods usually provide superficial and temporary respite. They often yield contradictory outcomes by further agitating the perpetrator, thus fueling hate, and even pose as restraints to online free speech, which leads to demonstrations. The steep growth of online hate speech is difficult for human regulators to keep up with, which is further aggravated by public access to large language models (LLMs) \cite{zhao2023survey} that are capable of generating fluent yet synthetic content at scale. Hence, there is a need for deep and scalable means for amicably addressing hate speech. Instead of only being reactive, a solution should aim at changing the perception of the initiator of hate speech. It should try to minimize the rhetoric by either persuading the perpetrator of hate speech contrary to their standpoint or at the least trying to broaden their perception and making them aware of the other facets of the topic of discussion. Online conversational agents can be such a robust solution to countering online hate.

Persuasive conversations that strive to change the mental state of an interlocutor rely on subtler human-based appeals. According to the ancient Greek philosopher Aristotle, persuading an audience depends on how well a person appeals in three areas: logos, ethos, and pathos \cite{rapp2002aristotle}. Recent research has engendered general-purpose LLMs like ChatGPT \cite{ouyang2022training}, GPT-4, Bard \cite{thoppilan2022lamda}, and inter-alia, whose fluent conversational capabilities have left the public in awe. Although impressive, without being explicitly trained to persuade, their efficacy in the purposeful and nuanced human task of persuasion is questionable. Hence, there is a need to develop controllable conversational systems that can incorporate such appeals while generating persuasive counter-hate responses. Here we attempt to consolidate and establish a set of features from diverse sources that can aid control and guide computational models to construct counter-arguments to hateful comments in online conversations. We broadly identify two feature categories for capturing the different aspects of a response: (i) Value-based: Features based on human qualities and characteristics such as Big-5 personality traits \cite{inbook, costa1992revised} and human values \cite{rokeach1973nature, https://doi.org/10.1111/j.1540-4560.1994.tb01196.x}. (ii) Structure-based: Features based on the patterns of reasoning such as Walton argument schemes \cite{walton2008argumentation} and argumentative discourse-based speech acts \cite{stolcke-etal-2000-dialogue} such as denouncing, questioning, etc. We incorporate a two-step process where first, we automatically annotate available hate speech dialogue datasets with the value and structure-based features using computational models. Second, using the annotated features as guides, we experiment with computational models for generating counter-arguments to hate speech. We experiment with different feature combinations to determine the best features for guiding counter-argument generation and report automatic and human evaluation results. In the process, we yield a silver-standard annotated version of the hate speech dialog corpora, with additional computational models for automatically annotating text with the defined features. We summarize our contributions as follows: 
\begin{itemize}
    \item We derive features from psychology, linguistics, philosophy, and social science and guide computational models for counter-hate argument generation.
    
    
    \item We release a computational framework for automatically identifying personality traits, human values, argument schemes, and argument types from argument text.

    \item We experiment and analyze the best combination of the features for response generation.
\end{itemize}

\noindent
Please note that this paper is strictly linguistically motivated and does not compose any psychographic or demographically motivated data profiling, analysis, or segmentation. The proposed methods only utilize diverse sources to derive features based on single turns of short text around specific topics, with the motivation being identifying observable stylometric differences between hateful and counter-hate response text. We use the features as control codes \cite{keskarCTRL2019} for aiding language modeling for counter-argument generation and hypothesize that such control codes can better partition the learned probability distribution. Furthermore, since the datasets used in the experiments are created synthetically by experts, the derived textual features do not portray the actual traits of users in a live setting. Through our experiments, we only highlight the differences in the presence of features, and the actual labels do not matter for our purpose.



    
    

\section{Related Work}

There has been noticeable advancement in gearing technology to curb the menace of online hate \cite{surv_kiritchenko2021confronting, surv_windisch2021protocol}. 
\citet{zhu2021generate} proposed a three-module pipeline approach where they generate multiple counter-arguments, and filters using a BERT-based model \cite{devlin2018bert}, and select the most relevant response using a novel retrieval-based method. \citet{saha2022countergedi} implemented CounterGeDi, an ensemble approach to enable DialoGPT \cite{zhang2019dialogpt} generate polite, detoxified, and emotionally laden counterspeech. \citet{de2021toxicbot} introduced Toxicbot, a system that detects and generates responses to intervene during online conversations with hate speech content. \cite{chung2021towards} introduced an external knowledge-based counter-narrative generation pipeline that can generate factual arguments across multiple domains. \citet{chung2020italian} leveraged GePpeTto for effectively generating counter-narratives in Italian.

Most available counter-hate speech corpora are limited to single-turn pairs of hate speech and counter-hate response, while very few focus on multi-turn dialogues. \citet{tekiroglu2020generating} presented a study on using GPT-2 \cite{radford2019language} for generating responses to hate and produced a silver-standard annotated dataset. \citet{conan} introduced Conan, which contains hate speech/counter-narrative pairs in English, French, and Italian about Islamophobia. Using a Human-in-the-Loop approach, \citet{multiconan} released Multi-Target Conan. They expanded Conan to include English examples across multiple hate targets. \citet{knowledgeconan} released a knowledge-grounded subset of Multi-Target Conan examples. \citet{contrahate} collected responses to the hateful comments in the Hateval corpus \cite{basile-etal-2019-semeval} and further enriched it with annotations based on Wagemanns periodic table of arguments \cite{wagemans2016constructing}. \citet{chasm} employed LLMs to create Chasm, a synthetic and quality-controlled counter-hate dataset. They provided further annotations that can also facilitate research in counter-narrative evaluation. \citet{dialoconan} shared DialoConan, a multi-turn synthetically generated and quality-controlled dialogue dataset between a hater and an NGO operator. We use DialoConan for our experiments and incorporate diverse features as control codes \cite{keskarCTRL2019} to better guide counter-argument generation.

\begin{table*}[!t]
\centering
\resizebox{\textwidth}{!}{%
\begin{tabular}{l|l|l|l}
\hline
\multicolumn{1}{l|}{\textbf{Type}} &
  \multicolumn{1}{l|}{\textbf{Source}} &
  \multicolumn{1}{l|}{\textbf{Derived From}} &
  \multicolumn{1}{c}{\textbf{Features / Control Codes}} \\ \hline
\multirow{2}{*}{Value} &
  \begin{tabular}[c]{@{}l@{}}Big-5 \\ personality\end{tabular} &
  \begin{tabular}[c]{@{}l@{}}Psychology \& \\ Social Science\end{tabular} &
  \begin{tabular}[c]{@{}l@{}}Openness, Conscientiousness, Extraversion, \\ Agreeableness, Neuroticism\end{tabular} \\ \cline{2-4} 
 &
  \begin{tabular}[c]{@{}l@{}}Human \\ Values\end{tabular} &
  \begin{tabular}[c]{@{}l@{}}Psychology \& \\ Social Science\end{tabular} &
  \begin{tabular}[c]{@{}l@{}}Achievement, Benevolence: caring, Security: personal, Security: \\ societal, Self-direction: action, Universalism: concern\end{tabular} \\ \hline
\multirow{2}{*}{Structure} &
  \begin{tabular}[c]{@{}l@{}}Argument \\ Scheme\end{tabular} &
  \begin{tabular}[c]{@{}l@{}}Philosophy \&\\ Linguistics\end{tabular} &
  \begin{tabular}[c]{@{}l@{}}From consequence, From source authority/knowledge, \\ Goal from means/means for goal, Rule or principle\end{tabular} \\ \cline{2-4} 
 &
  \begin{tabular}[c]{@{}l@{}}Argument \\ Type\end{tabular} &
  \begin{tabular}[c]{@{}l@{}}Linguistics\\ Speech Acts\end{tabular} &
  Denouncing, Facts, Hypocrisy, Positive, Question \\ \hline
\end{tabular}%
}
\caption{Description of different types of features.}
\label{tab:feature-table}
\end{table*}

\section{Defining and Annotating Features}
\label{features}

As described in Table \ref{tab:feature-table}, we derive features from psychological and characteristic-based Big-5 personality traits \cite{sanford1942speech, saha-etal-2022-stylistic} and human values \cite{https://doi.org/10.1111/j.1540-4560.1994.tb01196.x, kiesel:2022b}, logic and reasoning-based argument schemes \cite{walton2008argumentation, kondo-etal-2021-bayesian} from philosophy, and linguistic speech act-based \cite{stolcke-etal-2000-dialogue} counter-argument types \cite{conan} and bucket them into two broad categories: Value and Structure, for our purpose. Overall we use 20 distinct features in our experiments and automatically annotate DialoConan with such features using computational models, which we discuss below.

\subsection{Big 5 Personality Traits}

Personality is the most fundamental dimension of variation between humans, and it greatly influences our situational reactions, thoughts, feelings, expressions, and actions \cite{mairesse2007using}. We use the Big-5 personality trait \cite{inbook, costa1992revised} classifier released by \citet{saha-etal-2022-stylistic} to identify the most likely personality traits of the proponent of a hate speech text. We use the parser to parse each turn independently and annotate the hate and counter-hate argument with the Big-5 traits. Figure \ref{fig:big5} (Appendix \ref{sec:appendix}) illustrates the distribution of each trait by the type of speech. We observe hateful comments to be classified as more neurotic, whereas counter-hate responses contain higher proportions of the other four classes.

\subsection{Human Values}

Human values \cite{rokeach1973nature, https://doi.org/10.1111/j.1540-4560.1994.tb01196.x} are a set of beliefs that condition human being and influences their actions in different scenarios. It is a system based on cultural, social, and personal factors and varies across persons. Although they are well-researched in social science and already studied in formal argumentation settings \cite{bench2003persuasion}, it is only recently that computational models \cite{kiesel:2022b} have been used to identify them from arguments. \citet{kiesel:2022b} defined 20 human value categories (L2 categories) that capture four broad aspects (L3 categories) of values: openness to change, self-enhancement, conservation, and self-transcendence. Each L2 category encapsulates finer distinct human values (L1 category) that better characterizes each L2 category. For example, the L2 category of `Self-direction: thought' represents the values `Be creative', `Be curious' and `Have freedom of thought'. Furthermore, each value is described by a set of sentences (value descriptors, 218 in total) explaining what it means to possess such a human value. For example, `promoting imagination' and `being more creative' are two of the several descriptors for the value `Be creative`. We train computational models to automatically identify the L2 values from the hateful and counter-hateful text and use them as value-based features. We use the standard training and validation splits from the SemEval shared task by \citet{mirzakhmedova:2023a} and include the samples from the validation-zhihu split in the training samples, which are derived from the Chinese question-answering website Zhihu \footnote{https://www.zhihu.com/explore}. Due to class imbalances, we limit ourselves to the top 6 value categories: Achievement, Benevolence: caring, Security: personal, Security: societal, Self-direction: action, and Universalism: concern out of the twenty categories defined by Matt. Inspired by \citet{saha2023rudolf}, we implement a majority-based ensemble approach comprising three models, which we discuss below.

\subsubsection{Classification Based Model}
We fine-tune Roberta large \cite{liu2019roberta} with a multi-task objective of predicting L1, L2, and L3 labels from arguments. We encode each example using Roberta's pooler representation, and for each level, we use a linear layer to yield the logits. The model is optimized end-to-end by minimizing a weighted averaged Binary Cross-Entropy (BCE) loss function, with 0.23, 0.33, and 0.44 as the respective weights. We use AdamW \cite{adamw}  optimizer with a learning rate of 1e-5, and train until the validation loss stops decreasing for 4 consecutive epochs. We use the standard training and validation splits and additionally merge the validation-zhihu split in the training samples, yielding 5,493 training and 1,896 validation examples.

\subsubsection{Entailment Based Model}
We implement an entailment-based model for identifying the L1 value descriptors that an argument text entails. We transform the dataset conducive for textual entailment by creating positive and negative argument and value descriptor pairs and aggregate the predictions at the L2 level. We follow the same approach as \citet{saha2023rudolf} to create the entailment pairs. We also merge the validation-zhihu split with the training samples, resulting in 189,312 training and 65,900 validation examples. The entailment model comprises a Roberta-base encoder and a linear layer to predict the logit. During training, we concatenate an argument and value descriptor text and pass the encoder pooled representation through a linear layer to predict the logit. Similar to \citet{saha2023rudolf}, we load the model weights with pre-trained MNLI weights and fine-tune by minimizing the BCE loss function using AdamW optimizer and a learning rate of 1e-5.

\subsubsection{Similarity Based Model}
We implement an embedding similarity-based approach for determining which of the 218 value descriptors an argument text most resonates with. First, we train a metric learning-based model to learn embeddings of the 218 L1 value descriptors. Next, treating the descriptor embeddings as fixed centroids, we train an embedder that transforms and embeds argument text to the embedding space of the centroids. Finally, we determine the best descriptor of an argument text by computing the cosine similarity between its embedding and the descriptor centroids and aggregate the predictions at the L2 level.

\noindent
\textbf{Step 1: Generate human value descriptor embeddings:}
We create multiple training quadruples per value descriptor, consisting of the descriptor as the anchor element, one positive, and two negative descriptors. The positive descriptor belongs to the same L1 class as the anchor. For the negative descriptors, we randomly sample one descriptor from the same L2 but different L1 class and another from a different L2 and L1 class altogether, which we term as hard and easy negatives. Overall we create 702 examples and randomly use 90\% examples for training, and remaining for validation. 

We learn embeddings using a Roberta-based encoder by optimizing a metric learning-based loss function. The training objective is two folds: (i) Minimizing the distance between the anchor and the positive descriptor embedding while maximizing the distance between the anchor and the easy-negative descriptor embedding, which enables descriptors of the same aspects (L1) of human values (L2) to be closer in the embedding space than dissimilar ones. (ii) Minimizing the distance between the hard-negative and anchor descriptor while maximizing the distance between the hard-negative and the positive descriptor embedding, which enables descriptors of the different aspects (L1) of the same human values (L2) to be close to the anchor while still being far enough in comparison to the descriptors of the same aspects in the embedding space. The below set of equations formally details the learning algorithm.
\begin{flalign}
&\mathrm{enc(x)  = \mathrm{Roberta\_base\_pooler}(x)} &\\
&\mathrm{a = enc(anchor) \ ;\  p = enc(positive)}&\\
&\mathrm{n_x = enc(x\ negative)\mathrm{, x}\in \{easy, hard\} }&\\
&\mathrm{D(x,y) = cosine\_similarity(x,y)} &\\
&L \mathrm{= \alpha * [D(a,p) - D(a,n_{easy})] + }&&\\\nonumber &\ \ \ \ \ \ \ \ \beta * \mathrm{[D(n_{hard},a )-D(n_{hard},p)] + M }
\end{flalign}
Where M is a fixed margin that is set to 1.0, and $\alpha$=2.0, $\beta$=1.0, and are chosen empirically. We optimize the model using AdamW optimizer with a learning rate of 2e-5, and train until the validation loss stops decreasing for 5 consecutive epochs.

\noindent
\textbf{Step 2: Train argument embedder:}
We train an argument embedder for transforming and embedding an argument text to the embedding space of the value descriptors. For every example, we create multiple positive and negative argument-value descriptor pairs. If there are K-positive descriptors per argument, we ensure a balanced negative descriptor by randomly sampling one value descriptor per L1 class and persisting K random values. The training and validation dataset comprises 200,059 and 69,607 argument-value descriptor pairs. We pass the Roberta pooler representation of an argument text through 3 fully-connected layers with ReLU activation and train by minimizing the cosine distance between its embedding and positive value descriptor. We train the model end-to-end and use the same training parameters as the value descriptor embedder. During inference, we determine the best descriptor of an argument text by computing the cosine similarity between its embedding and the descriptor centroids and aggregate the predictions at the L2 level.

\subsubsection{Results}

Table \ref{tab:hum-val-results} shares the results for each of the three models, along with a majority-based ensemble of the three models. We report the macro-averaged F1 score for the positive class and overall level for all the twenty and top 6 human value classes. Overall the ensemble approach almost always yields better results. The classification-based model has the best validation results of the three individual models, which we attribute to its more parameters. We use the ensemble for automatically annotating each turn's hate and counter-hate text of the target dataset with the top 6 human values. Figure \ref{fig:hval} (Appendix \ref{sec:appendix}) plots the distribution of each value. We observe hateful comments to exhibit significantly more societal security concerns, whereas counter-hate responses manifest higher personal security and care.

\begin{table}[h]
\centering
\resizebox{\columnwidth}{!}{%
\begin{tabular}{l|cc|cc}
\hline
\multicolumn{1}{c|}{\textbf{}} &
  \multicolumn{2}{c|}{\textbf{All Classes}} &
  \multicolumn{2}{c}{\textbf{Top Classes}} \\ \cline{2-5} 
\multicolumn{1}{l|}{\textbf{Model}} &
  \multicolumn{1}{c}{\textbf{Positive}} &
  \multicolumn{1}{c|}{\textbf{Macro}} &
  \multicolumn{1}{c}{\textbf{Positive}} &
  \multicolumn{1}{c}{\textbf{Macro}} \\ \hline
Classification & 0.48          & 0.70          & 0.60          & \textbf{0.73} \\
Entailment                             & 0.47          & 0.66          & 0.61          & 0.68          \\
Similarity                             & 0.44          & 0.64          & 0.56          & 0.63          \\
Ensemble       & \textbf{0.52} & \textbf{0.71} & \textbf{0.63} & 0.72          \\ \hline
\end{tabular}%
}
\caption{F1 scores on validation set for Human Value Detection from text.}
\label{tab:hum-val-results}
\end{table}

\subsection{Argument Schemes}

Argument schemes are typical reasoning and inference patterns found in arguments. Walton provided an in-depth study of argument schemes \cite{walton2008argumentation} and defined 60 such schemes prevalent in daily argument text. \citet{kondo-etal-2021-bayesian} defined Bayesian networks to represent the inference pattern of arguments, which inspired \citet{saha2023argu} to experiment with computational models to detect the type of reasoning used by an argument text. They restricted to 6 of Walton's argument schemes: ``Means for Goal'', ``Goal from Means'', ``From Consequence'', ``Source Knowledge'', ``Source Authority'', and ``Rule or Principle''. We use their computational model to identify the most likely scheme used by each turn's hate and counter-hate argument. Based on the distribution of the labels, we further combine the classes ``From Source Authority'' with ``From Source Knowledge'' and ``Goal for Means'' with ``Means for Goal''. Figure \ref{fig:scheme} (Appendix \ref{sec:appendix}) plots the distribution of each scheme. We observe that counter-hate arguments are significantly more authoritative and knowledge-based. 

\begin{table*}[t]
\centering
\resizebox{0.8\textwidth}{!}{%
\begin{tabular}{l|l|cccccc}
\hline
\multicolumn{1}{l|}{\textbf{Model}} &
  \multicolumn{1}{l|}{\textbf{Variant}} &
  \textbf{Denouncing} &
  \textbf{Facts} &
  \textbf{Humor} &
  \textbf{Hypocrisy} &
  \textbf{Positive} &
  \textbf{Question} \\ \hline
Roberta & Masked     & 0.80          & \textbf{0.79} & 0.70          & \textbf{0.76} & \textbf{0.67} & \textbf{0.86} \\
Roberta & Non-masked & 0.76          & 0.77          & \textbf{0.81} & 0.69          & 0.62          & 0.80          \\ \hline
Deberta & Masked     & 0.76          & \textbf{0.79} & 0.73          & 0.71          & \textbf{0.67} & 0.83          \\
Deberta & Non-masked & \textbf{0.81} & 0.78          & 0.80          & 0.72          & 0.66          & 0.85          \\ \hline
\end{tabular}%
}
\caption{Macro-F1 scores on validation set for detecting counter-argument type from text.}
\label{tab:arg-valid-self}
\end{table*}

\begin{table*}[t]
\centering
\resizebox{0.9\textwidth}{!}{%
\begin{tabular}{l|l|ccccc|c}
\hline
\multicolumn{1}{l|}{\textbf{Model}} &
  \multicolumn{1}{l|}{\textbf{Variant}} &
  \textbf{Denouncing} &
  \textbf{Facts} &
  \textbf{Humor} &
  \textbf{Hypocrisy} &
  \textbf{Question} &
  \textbf{Average} \\ \hline
Roberta & Masked      & 0.45          & 0.79 & 0.53          & 0.38 & \textbf{0.80} & 0.59 \\
Roberta & Non-masked  & 0.61          & 0.78          & \textbf{0.68} & \textbf{0.43} & 0.74          & \textbf{0.65} \\ \hline
Deberta & Masked      & 0.46          & 0.78 & 0.46          & 0.33          & 0.69 & 0.54          \\
Deberta & Non-masked  & 0.47 & 0.74          & 0.55          & 0.33          & 0.71          & 0.56          \\ \hline
\cite{chung-etal-2021-multilingual}        & Monolingual & 0.65          & 0.84          & 0.45          & 0.35          & 0.73          & 0.60          \\
\cite{chung-etal-2021-multilingual}        & Translated  & \textbf{0.70} & \textbf{0.85} & 0.56          & 0.40          & 0.75          & \textbf{0.65} \\ \hline
\end{tabular}%
}
\caption{Macro-F1 scores on external test set for detecting counter-argument type from text.}
\label{tab:arg-valid-multiCN}
\end{table*}

\subsection{Argument Type}
Identifying the speech act of an argument can be crucial and provide more information about what kind of arguments generally work in a hateful scenario. There are scenarios in toxic discussions where responding with a fact might be more effective in reducing the rhetoric than just denouncing and challenging the standpoint of the hateful argument. For example, the comment 'immigrants are eating into the American job market and should be banned' can be countered with a fact around how immigrants only take up those jobs that Americans don't, instead of just denouncing it. \citet{conan} identified and defined ten categories of counter-responses to Islamophobic hateful comments: affiliation, consequences, denouncing, facts, humor, hypocrisy, positive, question, negative, and others. Due to the skewed class balance, we limit to the first 6 categories and experiment with computational models for automatically identifying such acts from counter-hate arguments. 

\subsubsection{Model Architecture}
We experiment with pre-trained Roberta \cite{liu2019roberta} and Deberta \cite{he2021deberta} large encoder-based transformer architectures \cite{vaswani2017attention} and expand the encoder's embedding layer with 6 tokens representing each counter-argument type. The encoder input a hate and counter-hate comment-response pair enclosed by beginning (BOS) and end-of-string (EOS) tokens and separated using two separator tokens. The encoder input is prefixed with the 6 tokens to facilitate the learning representation of each counter-argument type. We pass the encoder representation for the argument type and the BOS tokens through two multi-headed attention layers with four attention heads. We independently apply a linear layer to the resultant encoding of each argument type and predict its logit. The model is trained end-to-end by minimizing the BCE loss function using AdamW optimizer and a learning rate of 1e-5.

Since the Conan dataset only pertains to Islamophobia, we employ a strategy to mask Islamophobic keywords to make the model generalizable to other topics. We computationally curate a set of keywords for each discussion topic in the target dataset. We parse each sentence using Spacy \footnote{https://spacy.io} and preserve adjectives, adverbs, interjections, nouns, and verbs that are only present in a single category at most five times as category-specific frequent keywords of a topic. We further expand the keywords using Wordnet by adding related forms and pertainyms and also include their plural forms using inflect \footnote{https://github.com/jaraco/inflect}. We mask the keywords using a \textit{\#MASK\#} token and train a masked model variant. 

\subsubsection{Experiments and Results}

We use the English split of the Conan dataset for our experiments and include the French/Italian to English translated and paraphrased versions to yield 6,645 examples. We create a train/test set of 6,445/200 instances while ensuring that the counter-arguments present in the test set are unseen in the training set. We experiment with four model variants: masked and non-masked versions of Roberta and Deberta-large and report results in Table \ref{tab:arg-valid-self}. We highlight the best scores for each class in bold and observe that (i) The Roberta-based model usually outperforms Deberta with masking and vice-versa without masking. (ii) Except for the classes Denouncing and Humor, masking the inputs yields better results than non-masking. During inference, we create a majority-based ensemble of all four models and only annotate the counter-hate argument in each dialogue turn. We observe a low distribution of examples belonging to Humor and restrict ourselves to the top 5 classes.

To further validate the efficacy of our implementation, we re-train all four model variants on the train/test split by \citet{chung-etal-2021-multilingual} and compare our results in Table \ref{tab:arg-valid-multiCN}. We highlight the best scores for each class in bold and observe that (i) The Roberta-based model usually outperforms Deberta with and without masking. (ii) Except for the classes Facts and Questions, not masking the inputs yields better results than masking. (iii) Although we achieve the same overall score, our implementation achieves a better score for detecting Humor, Hypocrisy, and Questions. We attribute the differences in our model performance in Tables \ref{tab:arg-valid-self} and \ref{tab:arg-valid-multiCN} to the differences in data curation and size. \citet{chung-etal-2021-multilingual} filters out a significant amount of samples and only considers examples that belong to one of the top 5 classes. On the contrary, instead of filtering the data points, we task them as a multi-class binary prediction problem over the 6 most frequent classes.

\section{Response Generation}
We experiment with controllable computational models for generating responses in hateful dialogues. Our objectives are: (i) Assessing the feasibility and utility of using the value and structure-based features for generating and controlling the counter-arguments to hateful speech. (ii) Determining the best feature combination for optimal control over response generation. (iii) Understanding the importance of each of the four sets of features.

\subsection{Dataset}
We use the DialoConan dataset \cite{dialoconan}, which comprises over 3,000 fictitious multi-turn dialogues between a hater and an NGO operator, covering 6 targets of hate: LGBT+, migrants, Muslims, Jews, people of color, and women, and additionally include the hate and counter-hate pairs from the Conan dataset \cite{conan}. We use the computational models discussed in Section \ref{features} to automatically annotate the dialogue corpora, yielding a silver standard dataset.\footnote{The annotated datasets are strictly used only for research.}

\subsection{Model and Training}
\begin{figure}[h]
  \centering
  \includegraphics[width=\columnwidth]{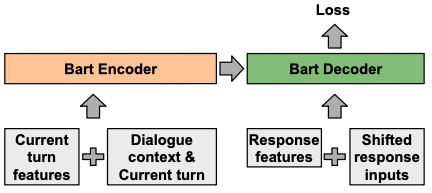}
  \caption{Response Generator Architecture.}
  \label{fig:generator}
\end{figure}

Illustrated in Figure \ref{fig:generator}, we use the transformer encoder-decoder-based Bart-base \cite{lewis-etal-2020-bart} as the response generator and expand its embedding layer to add the features as tokens. We concatenate the dialogue context and the current turn's hate speech (query) as the encoder input and additionally prefix the encoder input with the features only present in the current query text and not in the response. The decoder inputs comprise features that are only present in the current turn response and not in the query text, along with shifted response tokens. The model is trained end-to-end by minimizing the Cross-Entropy loss function using AdamW optimizer and a learning rate of 1e-5. During inference, the encoder input representation is kept similar to training, while the decoder inputs the desired response features and auto-regressively generates a response using beam search with a beam width of 5.


\subsection{Results and Analysis}
We perform extensive automatic and human evaluations to ascertain the usefulness of each category of features and report results.
\subsubsection{Automatic Evaluations}
To determine the best features, we train computational models with different combinations of the value (Val) and structure-based (Struct) features and compare automatic metrics such as corpus BLEU \cite{papineni-etal-2002-bleu}, Rouge-L \cite{lin-2004-rouge}, and language modeling perplexity (PPL). We train a model variant without any features as the internal baseline and report our results in Table \ref{tab:generation-auto-results}. For each depth of combination (1/2/3 features together) and metric, we highlight (in bold) the best-performing set of features and underline the overall best-performing feature combination per metric. 

\begin{table}[h]
\centering
\resizebox{\columnwidth}{!}{%
\begin{tabular}{c|l|l|ccc}
\hline
\textbf{ID} & \textbf{Type}           & \textbf{Features}                                                & \textbf{BLEU} & \textbf{RougeL} & \textbf{PPL}        \\ \hline
1& Baseline                & None                                                             & 7.83          & 23.18           & 7.88                \\ \hline 
2&\multirow{2}{*}{Val}    & big5                                                             & \textbf{8.43} & 23.76           & 7.79                \\
3&                        & humVal                                                           & 7.95          & 23.12           & 7.89                \\ \cline{2-6} 
4&\multirow{2}{*}{Struct} & argSch                                                           & 8.25          & \textbf{23.77}  & 7.88                \\ 
5&                        & argType                                                          & 8.10          & 23.37           & {\underline{ \textbf{7.65}}} \\ \hline
6& Val                     & humVal+big5                                                      & 7.37          & 22.72           & 7.85                \\ \cline{2-6} 
7& Struct                  & argSch+argType                                                   & 8.24          & 23.53           & \textbf{7.68}       \\ \cline{2-6} 
8& \multirow{4}{*}{\begin{tabular}[c]{@{}l@{}}Val+\\ Struct\end{tabular}} &
  argSch+big5 &
  {\underline{ \textbf{8.89}}} &
  {\underline{ \textbf{24.60}}} &
  7.83 \\
9&                        & humVal+argSch                                                    & 8.10          & 23.92           & 7.86                \\
10&                        & humVal+argType                                                   & 7.34          & 21.95           & 7.72                \\
11&                        & big5+argType                                                     & 7.15          & 22.31           & 7.71                \\ \hline
12&\multirow{5}{*}{\begin{tabular}[c]{@{}l@{}}Val+\\ Struct\end{tabular}} &
  \begin{tabular}[c]{@{}l@{}}humVal+big5\\ +argSch\end{tabular} &
  \textbf{7.52} &
  \textbf{23.13} &
  7.86 \\
13&                        & \begin{tabular}[c]{@{}l@{}}humVal+big5\\ +argType\end{tabular}   & 7.17          & 22.41           & 7.69                \\
14&                        & \begin{tabular}[c]{@{}l@{}}big5+argSch\\ +argType\end{tabular}   & 7.12          & 22.07           & 7.68                \\
15&                        & \begin{tabular}[c]{@{}l@{}}humVal+argSch\\ +argType\end{tabular} & 6.86          & 21.78           & 7.77                \\
16&                        & All                                                              & 6.75          & 22.23           & \textbf{7.66}       \\ \hline
\end{tabular}%
}
\caption{Automatic Evaluation Results of Generated Responses for All Feature Combinations.}
\label{tab:generation-auto-results}
\end{table}

\begin{figure}[h]
  \centering
  \includegraphics[width=\columnwidth]{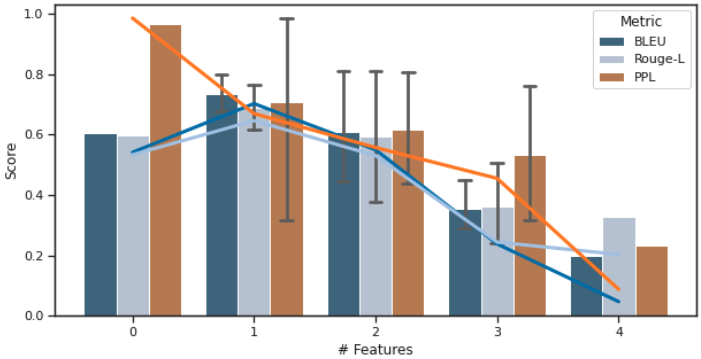}
  \caption{Comparison of Number of features and Normalized (between 0.2 \& 1) Score for Different Metrics.}
  \label{fig:num_feat_score}
\end{figure}

We observe that (i) When considered independently using the value-based Big-5 personality-based features yields the best BLEU scores, followed by the structure-based argument schemes, and vice-versa for the Rouge-L score. (ii) Considering feature pairs, combining the Big-5 personality and argument schemes produces the best paired and global results. (iii) Among the value-based features, using the Big-5 personality traits generally yields better results than human values. (iv) Among the structure-based features, using the argument scheme generally yields better results than the argument type. (v) Structure-based features generally yield better results than value-based features. However, using both types of features in tandem yields the best results (row 8). (vi) As depicted in Figure \ref{fig:num_feat_score}, combining three features decreases both BLEU and Rouge-L scores below baseline, and using all features together yields the worst BLEU score. Interestingly, the perplexity score generally improves (decreases) with more additional features. We reason that increasing the number of features better partitions the conditional probabilities of the tokens, which enhances language modeling, resulting in a lower perplexity score. However, it also increases the model complexity by adding more degrees of freedom and can be circumvented by additional training examples. We intend to validate this hypothesis as a future research step.

\subsubsection{Human Evaluations}
Employing two evaluators per sample in Amazon Mechanical Turk (AMT), we perform human evaluations on the variants that yield better BLEU scores than the baseline and report our results in Table \ref{tab:generation-result-humeval}. We define the following three metrics and holistically rate each argument text: (i) \textbf{Arg}umentativeness (higher better): On a Likert scale of 1 (low) to 5 (high), this indicates the cogency, novelty, and level of engagement of the generated argument. Incorporating proper reasoning, facts, opinions, and anecdotes, generally makes for a good, convincing, and engaging argument. (ii) \textbf{Flu}ency (higher better): On a binary scale (0=False, 1=True), this indicates if the generated response is grammatically and syntactically correct and penalizes violations of the rules of the English language other than spelling and punctuation mistakes. (iii) \textbf{Hal}lucination (lower better): On a binary flag (0=False, 1=True), this indicates if the generated argument is factually and logically sound. It penalizes arguments containing hallucinated facts or illogical propositions which commonly do not make sense in the real world. Figures \ref{fig:argumentativenes} and \ref{fig:hallucination} (Appendix \ref{sec:appendix}) illustrates the evaluation task in AMT.

\begin{table}[h]
\centering
\resizebox{\columnwidth}{!}{%
\begin{tabular}{c|l|l|ccc}
\hline
\textbf{ID} & \textbf{Type}                                                          & \textbf{Features} & \textbf{Arg}  & \textbf{Flu} & \textbf{Hal}        \\ \hline
1& Baseline                & None           & 3.80                & 0.98                & 0.06                \\ \hline
2& \multirow{2}{*}{Val}    & big5           & {\underline{ \textbf{3.85}}} & 0.98                & 0.06                \\
3&                        & humVal         & 3.67                & 0.99                & 0.06                \\ \cline{2-6} 
4& \multirow{2}{*}{Struct} & argSch         & 3.54                & 0.99                & 0.05                \\
5&                        & argType        & 3.78                & {\underline{ \textbf{1.00}}} & \textbf{0.04}       \\ \hline
6& Struct                  & argSch+argType & 3.67                & 0.98                & {\underline{ \textbf{0.03}}} \\ \cline{2-6} 
7& \multirow{3}{*}{\begin{tabular}[c]{@{}l@{}}Val+\\ Struct\end{tabular}} & argSch+big5       & \textbf{3.80} & 0.97         & {\underline{ \textbf{0.03}}} \\
8&                        & humVal+argSch  & 3.69                & \textbf{0.99}       & 0.04                \\
9&                        & All            & 3.58                & 0.98                & 0.09                \\ \hline
\end{tabular}%
}
\caption{Human Evaluation Results of Generated Responses for Feature Combinations Better than Baseline.}
\label{tab:generation-result-humeval}
\end{table}

For each depth of combination and metric in Table \ref{tab:generation-result-humeval}, we highlight (in bold) the best-performing set of features and underline the overall best-performing feature combination per metric. The annotators agree in 78\%, 90\%, and 97\% cases for argumentativeness, hallucinations, and fluency. We observe that (i) The responses from all model variants are fluent. (ii) For argumentativeness, models incorporating the personality-based features perform better than the baseline, whereas using only argument schemes rates the lowest. Combining personality and argument scheme-based features yields good results, whereas the responses from the model variant using all four features are rated low. (iii) For hallucinations, models incorporating the structure-based features perform better than the baseline and value-based features. Responses from models incorporating a combination of the structure and value-based features have a lower hallucination rate: Combining personality and argument scheme-based features yields good results. 

Overall, both automatic and human evaluations indicate that using Big-5 Personality and argument scheme-based features together performs best, demonstrating the need to use value and structure-based features for generating well-rounded arguments. Further, both types of evaluations indicate an upper bound to the optimal number of feature combinations, where responses using all four types of features are rated low. Table \ref{tab:generated-samples} (Appendix \ref{sec:appendix}) shares a few examples of the generated counter-hate arguments.

\section{Conclusion}
Motivated to minimize online hate by persuading perpetrators of hate speech contrary to their beliefs, we experiment with ways to generate controllable counter-hate arguments using features derived from psychology, philosophy, linguistics, and social science. We experiment with different feature combinations for generating fluent and persuasive counterarguments. We share computational models for automatically annotating text with such features and a silver-standard annotated hate speech dialog corpora. Our evaluations indicate that a convincing argument should generally appeal to the hateful perpetrator's personality while incorporating appropriate patterns of reasoning.

\section*{Acknowledgement}
We thank the anonymous reviewers for providing valuable feedback on our manuscript. This work is partly supported by NSF grant number IIS2214070. The content in this paper is solely the responsibility of the authors and does not necessarily represent the official views of the funding entity.

\section*{Limitations}
\begin{itemize}
    \item We automatically annotate all text with the identified features. The annotations are not validated by human evaluators.
    \item The hate-speech corpus used in the experiments mainly consists of synthetically curated dialogues. The argument generator's usefulness is not extensively validated on actual online hate comments.
    \item The hate-speech dialogue corpus used in the experiments is limited to a set of 6 topics in English. All model performance on any other topics or language is not validated.
\end{itemize}



\section*{Ethics Statement}
We acknowledge that all experiments were performed ethically and purely from an academic point of view. Although this research revolves around arguments from sensitive topics, the argument generators were not explicitly trained to be discriminatory, exhibit bias, or hurt anyone's sentiments. Further, any generated text does not reflect the stance of the authors. The human evaluators were appointed and compensated as per the legal norms of Amazon Mechanical Turk. 
This work is an attempt to improve on counterspeech generation ``in the lab", and is not intended for deployment. It is strictly linguistically motivated and does not compose any psychographic or demographically motivated data profiling, analysis, or segmentation. The proposed methods only utilize diverse sources to derive features based on single turns of short text around specific topics, with the motivation being identifying observable stylometric differences between hateful and counter-hate response text.

\bibliography{custom}
\bibliographystyle{acl_natbib}

\appendix

\section{Appendix}
\label{sec:appendix}

\begin{figure*}[h]
  \centering
  \includegraphics[width=\textwidth]{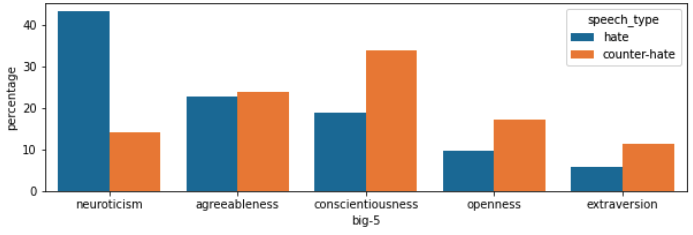}
  \caption{Distribution of Big-5 Personality Traits by type of speech.}
  \label{fig:big5}
\end{figure*}
\quad
\begin{figure*}[h]
  \centering
  \includegraphics[width=\textwidth]{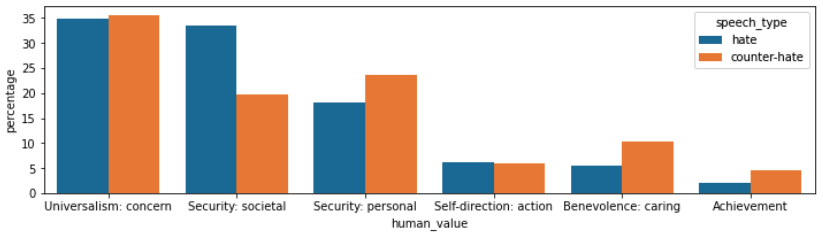}
  \caption{Distribution of Human Values by type of speech.}
  \label{fig:hval}
\end{figure*}
\quad
\begin{figure*}[h]
  \centering
  \includegraphics[width=\textwidth]{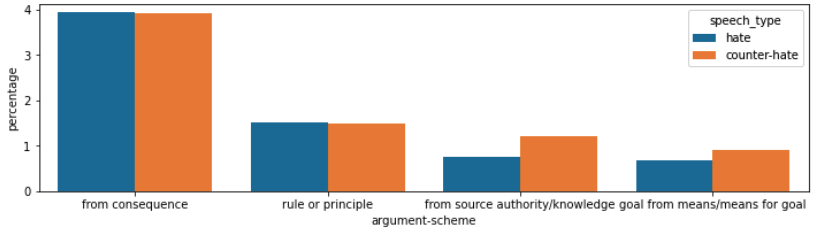}
  \caption{Distribution of Argument Scheme by type of speech. Note: The y-axis is scaled using cubic root.}
  \label{fig:scheme}
\end{figure*}

\begin{table*}[t]
\centering
\resizebox{\textwidth}{!}{%
\begin{tabular}{l|l|l|l|l}
\hline
\textbf{ID} &
  \multicolumn{1}{c|}{\textbf{Context}} &
  \multicolumn{1}{c|}{\textbf{Baseline}} &
  \multicolumn{1}{c|}{\textbf{All}} &
  \multicolumn{1}{c}{\textbf{argSch+big5}} \\ \hline
1 &
  \begin{tabular}[c]{@{}l@{}}\textbf{\textless{}hateSpeech\textgreater} the hoards of migrants that are \\ coming here are a threat to girls in our country, \\ migrant rape is an epidemic\end{tabular} &
  \begin{tabular}[c]{@{}l@{}}this is not true. how can \\ you say this about an \\ entire group of people?\end{tabular} &
  \begin{tabular}[c]{@{}l@{}}i don't see how rape can be co-\\ nsidered an epidemic when th-\\ ere are so many migrants com-\\ ing to our country to work and \\ provide for their families.\end{tabular} &
  \begin{tabular}[c]{@{}l@{}}rape is always a crime, \\ regardless of who the \\ rapist is. we should fo-\\ cus on stopping it inst-\\ ead of demonising it.\end{tabular} \\ \hline
2 &
  \begin{tabular}[c]{@{}l@{}}\textbf{\textless{}hateSpeech\textgreater} gays are perverts. the bible \\ teaches that homosexuals are to be despised, \\ a better world is possible only without them. \\ \textbf{\textless{}counterSpeech\textgreater} what is wrong with being gay \\ for you? i don't think it is wrong to love \\ someone of your same sex. \textbf{\textless{}hateSpeech\textgreater} gays \\ are different from other people because of their \\ sexual orientation. and deep down in their heart \\ they despise themselves.\end{tabular} &
  \begin{tabular}[c]{@{}l@{}}why do you say that? \\ do you have any evidence \\ for what you say?\end{tabular} &
  \begin{tabular}[c]{@{}l@{}}i don't understand why you \\ would say that. homosexuals \\ are not different from other \\ people, they just have \\ different sexual orientations.\end{tabular} &
  \begin{tabular}[c]{@{}l@{}}gay people don't \\ deserve to be despised \\ for who they are. they \\ are human beings with \\ the same rights as \\ anyone else.\end{tabular} \\ \hline
3 &
  \begin{tabular}[c]{@{}l@{}}\textbf{\textless{}hateSpeech\textgreater} jews are addicted to their religion \\ and should be isolated. \textbf{\textless{}counterSpeech\textgreater} even if \\ jews were addicted to their religion, there is noth-\\ ing wrong in them joining a community of other \\ people, regardless of the fact they're believers. \textbf{\textless{}h-}\\ \textbf{ateSpeech\textgreater} they can't join our community because \\ they don't belong to the united states of america.\end{tabular} &
  \begin{tabular}[c]{@{}l@{}}jews have been part of our \\ country since its creation \\ and have always had a place \\ here. they are our brothers \\ and sisters.\end{tabular} &
  \begin{tabular}[c]{@{}l@{}}jews have been part of our \\ country since its creation, and \\ have always had a place here. \\ they are our brothers and \\ sisters.\end{tabular} &
  \begin{tabular}[c]{@{}l@{}}the united states of \\ america is a melting \\ pot of cultures and re-\\ ligions. it is up to us to \\ make sure that jews fe-\\ el welcome here, rega-\\ rdless of their religion.\end{tabular} \\ \hline
4 &
  \begin{tabular}[c]{@{}l@{}}\textbf{\textless{}hateSpeech\textgreater} all these migrants are coming to \\ ruin our country, they are all rapists, drug addicts, \\ and don't have any right to be here. \textbf{\textless{}counterSpe-}\\ \textbf{ech\textgreater} migration is necessary, to bring in resources \\ and workers that can improve the lives of all. the \\ causes of addiction or sexual violence lie elsewh-\\ ere. \textbf{\textless{}hateSpeech\textgreater} they don't improve our lives, \\ immigrants break our laws! they are not welcome \\ here, they are not worth our time.\end{tabular} &
  \begin{tabular}[c]{@{}l@{}}immigrants come here \\ to work, pay taxes, and \\ have a better quality of life. \\ they are entitled to be here.\end{tabular} &
  \begin{tabular}[c]{@{}l@{}}they are not breaking our \\ laws, they are just trying to \\ live their life.\end{tabular} &
  \begin{tabular}[c]{@{}l@{}}immigrants come here \\ to work, pay taxes and \\ have a better life. they \\ are not thieves, they are \\ just trying to live their \\ life.\end{tabular} \\ \hline
\end{tabular}%
}
\caption{Samples of generated counter-hate arguments from the baseline generator, the best performing generator (argSch+big5), and the generator incorporating all four types of features (all). \textbf{OFFENSIVE CONTENT WARNING:} This table contains some examples of hateful content. This is strictly for the purposes of enabling this research, and we have sought to minimize the number of examples where possible. Please be aware that this content could be offensive and cause you distress.}
\label{tab:generated-samples}
\end{table*}

\begin{figure*}[h]
  \centering
  \includegraphics[width=\textwidth]{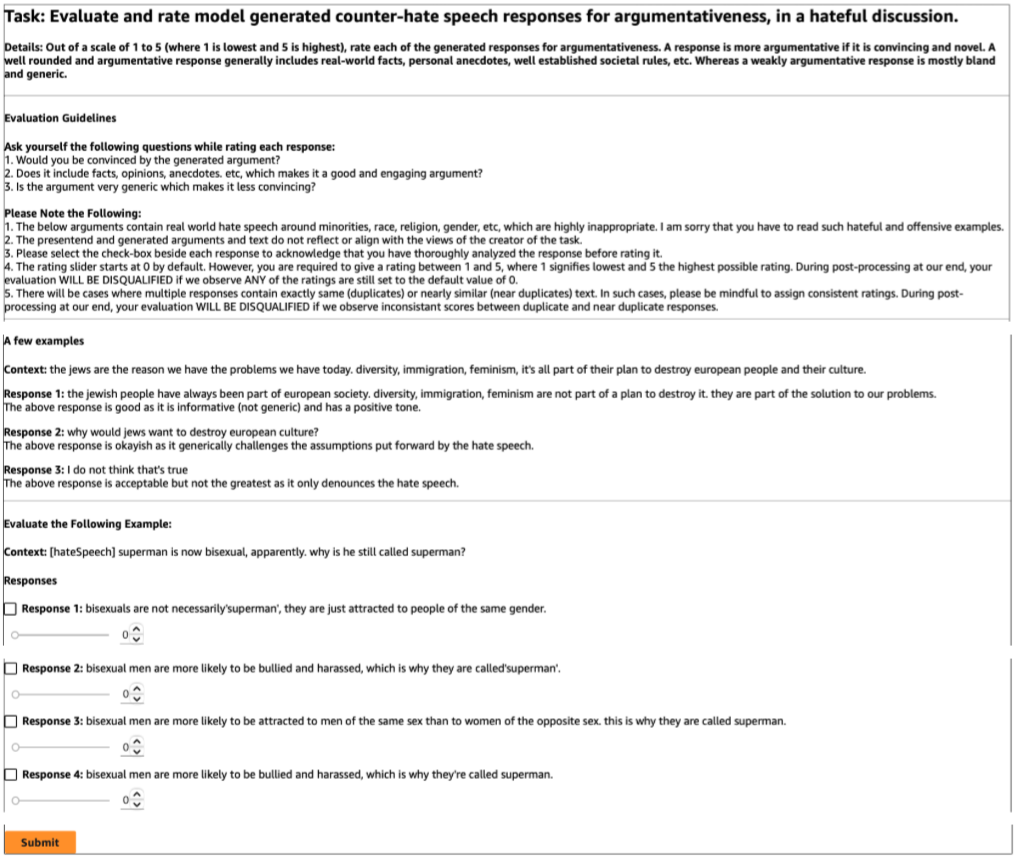}
  \caption{Screenshot of Amazon Mechanical Turk Human Intelligence Task for evaluating argumentativeness.}
  \label{fig:argumentativenes}
\end{figure*}

\begin{figure*}[h]
  \centering
  \includegraphics[width=\textwidth]{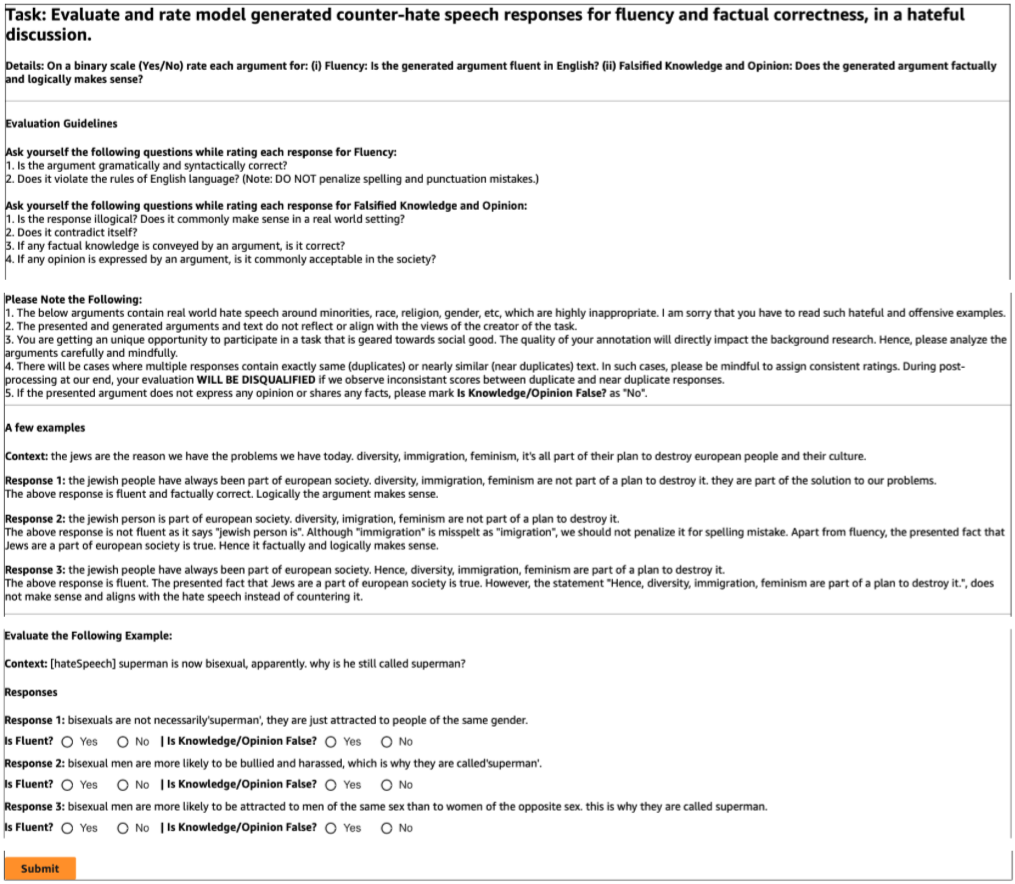}
  \caption{Screenshot of Amazon Mechanical Turk Human Intelligence Task for evaluating fluency and hallucination.}
  \label{fig:hallucination}
\end{figure*}

\end{document}